\pdfoutput=1

\documentclass[11pt]{article}

\usepackage[]{EMNLP2023}

\usepackage{amsmath, amssymb, amsfonts}

\usepackage{graphicx}
\usepackage{subcaption}
\usepackage{booktabs}
\usepackage{multirow}
\usepackage{colortbl}
\usepackage[table,xcdraw]{xcolor}
\usepackage{adjustbox}
\usepackage{tabularx}

\usepackage{algorithm}
\usepackage{algpseudocode}

\usepackage{times}
\usepackage{latexsym}
\usepackage[T1]{fontenc}
\usepackage[utf8]{inputenc}
\usepackage{microtype}
\usepackage{inconsolata}

\usepackage{cleveref}
\crefname{figure}{Fig.}{Figs.}
\crefname{algorithm}{Algorithm}{Algos.}

\usepackage{hyperref}

\definecolor{darkpink}{RGB}{200,0,120}

\hypersetup{
  colorlinks=true,
  linkcolor=black,
  citecolor=black,
  urlcolor=darkpink
}

\usepackage{fancyhdr}

\title{GateKD: Confidence-Gated Closed-Loop Distillation for Robust Reasoning}

\author{ Kasidit Sermsri\textsuperscript{\textdagger} \quad Teerapong Panboonyuen\textsuperscript{\textdagger,\textdaggerdbl,\S}\thanks{\hspace{0.5em}Corresponding author: Teerapong Panboonyuen. He conceived, designed, and led all major aspects of this work, including the research direction, methodology, implementation, experiments, analysis, and manuscript preparation. His contributions were central to the scientific rigor, technical innovation, and successful completion of the study. MARSAIL (Motor AI Recognition Solution Artificial Intelligence Laboratory) develops advanced AI solutions for the automotive and insurance industries. Project page available at: \href{https://kaopanboonyuen.github.io/GateKD}{https://kaopanboonyuen.github.io/GateKD}} \\ \textsuperscript{\textdagger}Chulalongkorn University \\ \textsuperscript{\textdaggerdbl}MARSAIL \\ 
\textsuperscript{\S}PBYAIL (Panboonyuen AI Lab) \\ \texttt{6532012521@student.chula.ac.th} \quad \texttt{teerapong.pa@chula.ac.th}  \\
  \\
   \href{https://kaopanboonyuen.github.io/GateKD}{https://kaopanboonyuen.github.io/GateKD}
}

\begin{document}
\maketitle

\begin{abstract}
Distilling multi-step reasoning abilities from large language models (LLMs) into compact student models remains challenging due to noisy rationales, hallucinated supervision, and static teacher–student interactions. Existing reasoning distillation methods, including mentor-based approaches, predominantly operate in an open-loop manner, implicitly assuming uniform teacher reliability and consequently propagating erroneous intermediate reasoning. We propose \textbf{GateKD}, a confidence-gated closed-loop distillation framework that enables robust reasoning transfer by treating the teacher as a dynamic gatekeeper rather than a static oracle. GateKD introduces three complementary mechanisms: (i) confidence-gated soft supervision that selectively distills reliable predictive signals, (ii) gated hidden-state evolution that aligns intermediate representations only when teacher confidence is high, and (iii) reliability-filtered attention distillation that preserves stable reasoning structures while suppressing noisy patterns. These components jointly form a closed feedback loop in which teacher confidence continuously modulates the distillation process, reducing hallucination transfer and stabilizing student reasoning. Extensive experiments across commonsense, logical, and symbolic reasoning benchmarks, using T5 and Flan-T5 backbones of varying sizes, demonstrate that GateKD consistently outperforms strong open-loop distillation baselines. Notably, GateKD yields substantial gains in logical and symbolic reasoning, remains robust under low-resource distillation settings, and shows clear performance degradation when any gating component is removed. Our results highlight that confidence-gated closed-loop supervision is critical for building reliable and scalable small reasoning models.
\end{abstract}

\section{Introduction}

Large language models (LLMs) have demonstrated remarkable multi-step reasoning abilities when guided to explicitly articulate intermediate reasoning steps, a capability formalized through Chain-of-Thought (CoT) prompting \citep{wei2022chain}. Subsequent studies show that reasoning reliability can be further enhanced by aggregating multiple reasoning trajectories, as in self-consistency decoding, which mitigates brittle or spurious inference paths \citep{wang2022selfconsistency}. Complementary to sampling-based approaches, structured prompting strategies such as least-to-most prompting decompose complex problems into simpler subproblems, highlighting the central role of intermediate reasoning structure in effective inference \citep{zhou2022leastmost}. Collectively, these advances suggest that \textbf{how} reasoning is constructed is as important as \textbf{what} prediction is produced.

Motivated by the high computational cost of LLMs, recent work has focused on transferring these reasoning capabilities to smaller and more efficient models via reasoning distillation. Early approaches distill soft labels or explicit reasoning traces from large teachers, demonstrating that students can acquire non-trivial reasoning skills despite limited capacity. Building on this line of research, Mentor-KD \citep{lee2024mentor} introduces a task-specific mentor model that augments both rationales and soft supervision, substantially improving multi-step reasoning distillation, particularly under low-resource settings. This work reveals a key insight: \textbf{intermediate supervision tailored to the task can be more effective than relying solely on large, generic LLM teachers}.

Despite these advances, existing reasoning distillation methods—including Mentor-KD—predominantly operate in an \emph{open-loop} fashion. Teacher or mentor signals are treated as uniformly reliable and transferred wholesale to the student. In practice, however, even strong teachers exhibit fluctuating confidence across inputs, producing hallucinated reasoning steps, unstable intermediate representations, or misleading attention patterns. Blindly distilling such signals risks amplifying noise, especially for small models that lack the capacity to recover from erroneous supervision. This observation raises a fundamental question: \emph{should all teacher reasoning be trusted equally during distillation?}

In this work, we answer this question negatively and propose \textbf{GateKD}, a confidence-gated closed-loop distillation framework for reasoning transfer. The key idea behind GateKD is simple yet powerful: instead of passively absorbing teacher signals, the student selectively learns from them based on the teacher’s predictive confidence. Specifically, GateKD estimates teacher reliability via predictive entropy and uses this signal to dynamically gate three complementary distillation pathways: \textbf{(i)} soft-label supervision, \textbf{(ii)} hidden-state alignment, and \textbf{(iii)} attention distillation. As a result, reliable reasoning patterns are reinforced, while uncertain or unstable signals are suppressed.

Our design is inspired by Mentor-KD \citep{lee2024mentor}, but departs from it in a crucial way. While Mentor-KD focuses on \emph{where} to source better supervision—by introducing a task-specialized mentor—GateKD focuses on \emph{when and how} supervision should be transferred. By incorporating confidence-aware gating, GateKD transforms reasoning distillation from a static, open-loop procedure into a closed-loop interaction between teacher and student. This perspective aligns with the intuition underlying self-consistency and structured prompting: robust reasoning does not arise from a single trajectory, but from selectively trusting stable and coherent inference paths.

Empirically, we demonstrate that GateKD consistently outperforms strong open-loop distillation baselines across commonsense, logical, and symbolic reasoning benchmarks, and across multiple student model scales. Notably, the gains are most pronounced on logical and symbolic tasks, where erroneous intermediate reasoning is particularly detrimental. These results indicate that \textbf{selective reasoning transfer}, rather than indiscriminate distillation, is critical for building reliable small-scale reasoners.

In summary, our contributions are threefold:
\textbf{(i)} we identify confidence misalignment as a key limitation of existing reasoning distillation methods;
\textbf{(ii)} we propose GateKD, a unified confidence-gated framework that selectively distills reliable reasoning signals; and
\textbf{(iii)} we empirically show that closed-loop, confidence-aware distillation yields robust and scalable reasoning improvements for small language models.

\section{Related Work}

\subsection{Reasoning in Large Language Models}

Large language models (LLMs) have demonstrated strong multi-step reasoning capabilities when prompted to explicitly generate intermediate reasoning steps, a phenomenon formalized as Chain-of-Thought (CoT) prompting \citep{wei2022chain}. Subsequent work shows that aggregating multiple reasoning trajectories via self-consistency decoding improves robustness by mitigating brittle or spurious inference paths \citep{wang2022selfconsistency}. Complementary to sampling-based approaches, structured prompting strategies such as least-to-most prompting decompose complex reasoning problems into simpler subproblems, highlighting the importance of intermediate reasoning structure in effective inference \citep{zhou2022leastmost}. These techniques collectively suggest that reasoning quality depends not only on model capacity, but also on the stability and coherence of intermediate reasoning processes. The strong reasoning abilities exhibited by large proprietary models such as GPT-4 further reinforce this observation \citep{openai2023gpt4}.

\subsection{Reasoning Distillation for Small Language Models}

Motivated by the computational cost of LLMs, a growing body of work explores distilling reasoning abilities into smaller student models. Early studies demonstrate that distilling soft labels or explicit reasoning traces enables small models to acquire non-trivial reasoning skills \citep{ho2023teaching,deng2023implicitcot}. Program-aided distillation (PaD) further shows that structured program supervision can outperform naive CoT fine-tuning for reasoning transfer \citep{zhu2023pad}. Mixed Distillation combines heterogeneous teacher signals, including logits and rationales, to improve student generalization \citep{li2024mixed}. Other works explore architectural or parameter-efficient strategies, such as mixture-of-experts distillation \citep{li2024route} and multilingual reasoning distillation \citep{payoungkhamdee2024multilingual}.

Several recent studies investigate how reasoning supervision should be structured. Cascading decomposed CoT distillation improves generalization by progressively transferring simpler reasoning steps \citep{dai2024improve}, while StepER performs step-wise distillation in retrieval-augmented reasoning settings \citep{lee2025steper}. CODI compresses explicit CoT into continuous representations via self-distillation, reducing inference cost while preserving reasoning behavior \citep{shenen2025codi}. Reinforcement-learning-based distillation methods further attempt to uncover implicit multi-branch reasoning structures within teachers \citep{xu2025rlkd}. Comprehensive analyses of CoT distillation identify key factors affecting reasoning transfer, including rationale quality, supervision granularity, and task difficulty \citep{chen2025unveiling-key,chen2025aclgeneral}.

\subsection{Mentor-Based and Multi-Path Distillation}

Mentor-based distillation has emerged as a promising direction for improving reasoning transfer. Mentor-KD introduces a task-specific mentor model to provide refined rationales and soft supervision, significantly improving multi-step reasoning distillation under limited data regimes \citep{lee2024mentor}. Related work explores learning from diverse reasoning paths via routing and collaboration mechanisms, emphasizing that not all reasoning trajectories are equally informative \citep{lei2025diversepaths}. These approaches highlight the importance of intermediate supervision quality, but still largely assume that provided teacher or mentor signals are uniformly reliable.

\subsection{Limitations of Open-Loop Distillation}

Despite their success, most existing reasoning distillation methods operate in an \emph{open-loop} fashion, treating teacher supervision as static and equally trustworthy across inputs. Empirical analyses show that even strong teachers frequently generate hallucinated or unstable reasoning steps, particularly for challenging logical and symbolic tasks \citep{hsieh2023rationale,song2023knowledge}. Blindly distilling such signals risks amplifying noise and degrading student robustness, especially for small models with limited capacity to recover from erroneous supervision.

\subsection{Our Position}

In contrast to prior work, \textbf{GateKD} introduces a confidence-gated closed-loop distillation paradigm that dynamically regulates \emph{when} and \emph{how} teacher reasoning signals are transferred. While Mentor-KD focuses on improving the \emph{source} of supervision via mentor models \citep{lee2024mentor}, GateKD focuses on selectively trusting supervision based on teacher confidence. By explicitly modeling supervision reliability and gating soft labels, hidden states, and attention patterns, GateKD complements existing reasoning distillation approaches and addresses a fundamental limitation of open-loop reasoning transfer.

\section{Approach}
\label{sec:approach}

We introduce \textbf{GateKD}, a \emph{confidence-gated closed-loop distillation framework} for robust multi-step reasoning transfer from large teacher models to compact student models.
Unlike prior reasoning distillation approaches that treat teacher supervision as uniformly reliable and static, GateKD dynamically modulates the influence of teacher signals based on their estimated reliability, forming an implicit closed feedback loop during training.

\begin{figure*}[t]
    \centering
    \includegraphics[width=\textwidth]{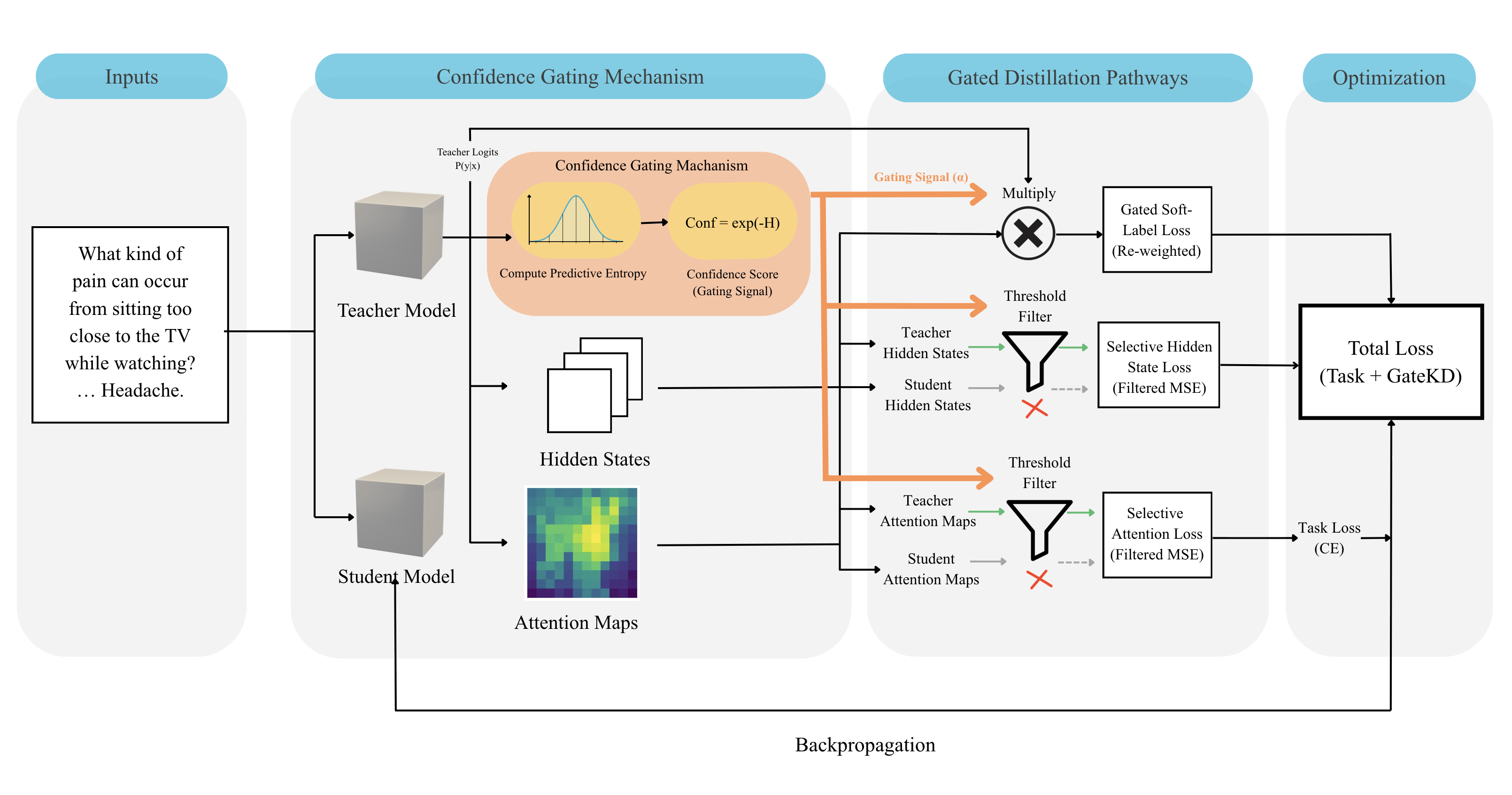}
    \caption{
    Overview of the proposed \textbf{GateKD} framework.
    Given an input, the teacher and student models process the instance in parallel.
    The teacher produces predictive distributions, hidden states, and attention maps, along with a confidence score estimated via predictive entropy.
    This confidence signal acts as a unified gating mechanism that selectively controls three distillation pathways:
    (i) \emph{confidence-gated soft supervision}, where teacher soft labels are re-weighted by confidence;
    (ii) \emph{gated hidden-state evolution}, where intermediate representations are aligned only when teacher confidence exceeds a threshold; and
    (iii) \emph{reliability-filtered attention distillation}, which transfers structural reasoning patterns selectively.
    All gated distillation losses are jointly optimized with the task loss, forming an implicit closed-loop interaction that suppresses unreliable teacher signals while preserving stable reasoning knowledge.
    }
    \label{fig:main_arch}
\end{figure*}

Figure~\ref{fig:main_arch} illustrates the overall architecture of GateKD.
Given an input instance, both teacher and student models process the input in parallel.
The teacher additionally produces a confidence signal that selectively gates which outputs, hidden representations, and attention structures are distilled to the student.
This design prevents hallucinated or unstable reasoning trajectories from being propagated, while preserving reliable reasoning knowledge.

\subsection{Overview of GateKD}

GateKD operates on three complementary distillation pathways:
\begin{enumerate}
    \item \textbf{Confidence-Gated Soft Supervision}, which adaptively reweights teacher predictive distributions;
    \item \textbf{Gated Hidden-State Evolution}, which selectively aligns intermediate reasoning representations; and
    \item \textbf{Reliability-Filtered Attention Distillation}, which transfers structural reasoning patterns only when reliable.
\end{enumerate}

All three pathways are jointly optimized alongside the standard task loss, allowing the student to learn accurate predictions while gradually acquiring stable reasoning behaviors.
Crucially, the gating mechanism is driven by teacher uncertainty estimated at training time, enabling dynamic modulation rather than fixed heuristics.

\subsection{Confidence Estimation and Gating Signal}

For each input $x$, the teacher model $\mathcal{T}$ produces a probability distribution
$p^{\mathcal{T}} = \mathrm{softmax}(z^{\mathcal{T}})$.
We estimate teacher confidence using predictive entropy:
\begin{equation}
C(x) = \exp\left(- \sum_i p^{\mathcal{T}}_i \log p^{\mathcal{T}}_i \right),
\end{equation}
where lower entropy corresponds to higher confidence.

This scalar confidence score serves as a \emph{unified gating signal} that modulates all distillation pathways.
Intuitively, when the teacher is uncertain, its supervision is downweighted or filtered out, preventing unreliable reasoning signals from influencing the student.

\subsection{Confidence-Gated Soft Supervision}

Standard knowledge distillation assumes that teacher predictions are equally informative for all training instances.
However, in reasoning-intensive tasks, teacher outputs may be unreliable due to spurious reasoning paths or hallucinated intermediate steps.

GateKD addresses this issue by weighting the soft-label distillation loss using the confidence signal:
\begin{equation}
\mathcal{L}_{\text{gate-soft}} =
\mathbb{E}_{x}\left[
C(x) \cdot
\mathrm{CE}\left(
p^{\mathcal{T}}, \log p^{\mathcal{S}}
\right)
\right],
\end{equation}
where $p^{\mathcal{S}}$ denotes the student’s predictive distribution.

This formulation softly suppresses low-confidence supervision rather than discarding it entirely, leading to smoother optimization and improved robustness compared to hard filtering strategies.

\subsection{Gated Hidden-State Evolution}

Beyond output-level supervision, intermediate hidden representations encode step-by-step reasoning trajectories.
Blindly distilling these representations can amplify teacher errors and destabilize training.

To mitigate this issue, GateKD selectively aligns hidden states only when the teacher is sufficiently confident.
Let $\phi(\cdot)$ denote a projection layer that maps teacher representations to the student’s hidden space.
The gated hidden-state loss is defined as:
\begin{equation}
\mathcal{L}_{\text{gate-hid}} =
\sum_{k=1}^{L_S}
\mathbb{I}\left[C_k > \bar{C}\right]
\left\|
h^{\mathcal{S}}_k - \phi\left(h^{\mathcal{T}}_{\alpha(k)}\right)
\right\|_2^2,
\end{equation}
where $\alpha(k)$ aligns student and teacher layers, and $\bar{C}$ is the batch-level mean confidence.

This selective alignment ensures that the student internalizes only stable reasoning representations, while allowing it to independently develop alternative representations when teacher uncertainty is high.

\subsection{Reliability-Filtered Attention Distillation}

Attention maps encode structural reasoning patterns such as dependency tracking and variable binding.
However, attention behavior is particularly sensitive to noisy or uncertain predictions.

GateKD applies the same confidence-based gating to attention alignment:
\begin{equation}
\mathcal{L}_{\text{gate-att}} =
\sum_{k}
\mathbb{I}\left[C_k > \bar{C}\right]
\left\|
A^{\mathcal{S}}_k - A^{\mathcal{T}}_{\alpha(k)}
\right\|_2^2.
\end{equation}

By distilling attention maps only when the teacher’s reasoning is reliable, GateKD preserves meaningful structural patterns while suppressing spurious attention behaviors.

\subsection{Overall Training Objective and Closed-Loop Interpretation}

The final training objective combines task supervision with gated distillation losses:
\begin{equation}
\mathcal{L} =
\mathcal{L}_{\text{task}} +
\lambda_1 \mathcal{L}_{\text{gate-soft}} +
\lambda_2 \mathcal{L}_{\text{gate-hid}} +
\lambda_3 \mathcal{L}_{\text{gate-att}}.
\end{equation}

Although GateKD does not explicitly retrain the teacher, the confidence-gated mechanisms induce an implicit \emph{closed-loop} interaction.
As the student improves over training, reliable teacher signals increasingly dominate the distillation process, while unreliable supervision is progressively suppressed.
This dynamic stands in contrast to open-loop distillation methods, where teacher guidance remains static throughout training.

Overall, GateKD enables robust and selective reasoning transfer, allowing compact student models to acquire stable multi-step reasoning capabilities without inheriting teacher hallucinations.

\section{Experiments}
We evaluate \textbf{GateKD} on a diverse set of reasoning benchmarks covering
\emph{commonsense}, \emph{logical}, and \emph{symbolic} reasoning.
Following prior work on reasoning distillation
\cite{lee2024mentor},
we adopt T5 and Flan-T5 model families as student backbones and employ
\texttt{large} variants as task-specific teachers.
All student models are trained using identical data splits and optimization settings
for fair comparison across distillation strategies.

\paragraph{Tasks.}
Commonsense reasoning is evaluated on CSQA and StrategyQA (SQA).
Logical reasoning is measured using the Shuffled Objects task,
and symbolic reasoning is assessed via the Last Letter concatenation task.
All metrics are reported in accuracy.

\paragraph{Baselines.}
We compare GateKD with several representative open-loop distillation methods,
including Vanilla-KD, MCC-KD \cite{lee2024mentor}, and Mentor-KD
\cite{lee2024mentor}.
In addition, we report zero-shot CoT performance of GPT-4o-mini as an upper-bound reference.
All student results are averaged over 5 random seeds.

\subsection{Main Results}
Tables~\ref{tab:t5_gatekd} and~\ref{tab:flant5_gatekd} summarize the main experimental results
across T5 and Flan-T5 backbones, respectively.
We observe that \textbf{GateKD consistently outperforms all open-loop distillation baselines}
across model scales and reasoning categories.

\begin{table*}[t]
\centering
\resizebox{\textwidth}{!}{%
\begin{tabular}{lclcccc}
\hline
Model & \#Params & Method &
\multicolumn{2}{c}{Commonsense} &
Logical &
Symbolic \\ \cline{4-7}
 &  &  &
CSQA & SQA &
Shuffled &
Last Letter \\ \hline
GPT-4o-mini & -- & ZS-CoT (teacher)
& 76.8 & 61.4 & 83.1 & 71.2 \\ \hline
T5-large & 780M & Vanilla-KD
& 69.3 $\pm$ 0.4 & 58.6 $\pm$ 0.6 & 88.1 $\pm$ 0.3 & 69.0 $\pm$ 0.5 \\ \hline
\multirow{4}{*}{T5-base} & \multirow{4}{*}{250M}
& Vanilla-KD
& 61.9 $\pm$ 0.5 & 55.2 $\pm$ 0.7 & 78.4 $\pm$ 0.6 & 56.1 $\pm$ 0.8 \\
& & MCC-KD
& 63.0 $\pm$ 0.6 & 56.4 $\pm$ 0.5 & 81.0 $\pm$ 0.7 & 58.2 $\pm$ 0.6 \\
& & Mentor-KD
& 64.2 $\pm$ 0.4 & 57.6 $\pm$ 0.6 & 84.9 $\pm$ 0.5 & 61.0 $\pm$ 0.7 \\
& & \cellcolor[gray]{.9}\textbf{GateKD (ours)}
& \cellcolor[gray]{.9}\textbf{66.8 $\pm$ 0.3}
& \cellcolor[gray]{.9}\textbf{59.9 $\pm$ 0.4}
& \cellcolor[gray]{.9}\textbf{90.6 $\pm$ 0.4}
& \cellcolor[gray]{.9}\textbf{65.7 $\pm$ 0.5} \\ \hline
\multirow{4}{*}{T5-small} & \multirow{4}{*}{80M}
& Vanilla-KD
& 55.4 $\pm$ 0.6 & 48.9 $\pm$ 0.8 & 63.7 $\pm$ 0.7 & 49.6 $\pm$ 0.9 \\
& & MCC-KD
& 56.8 $\pm$ 0.7 & 49.5 $\pm$ 0.6 & 66.1 $\pm$ 0.6 & 51.3 $\pm$ 0.7 \\
& & Mentor-KD
& 58.6 $\pm$ 0.5 & 51.8 $\pm$ 0.7 & 72.9 $\pm$ 0.5 & 55.2 $\pm$ 0.6 \\
& & \cellcolor[gray]{.9}\textbf{GateKD (ours)}
& \cellcolor[gray]{.9}\textbf{61.3 $\pm$ 0.4}
& \cellcolor[gray]{.9}\textbf{54.6 $\pm$ 0.5}
& \cellcolor[gray]{.9}\textbf{80.8 $\pm$ 0.4}
& \cellcolor[gray]{.9}\textbf{60.1 $\pm$ 0.5} \\ \hline
\end{tabular}}
\caption{
Performance comparison on commonsense, logical, and symbolic reasoning benchmarks using T5 backbones.
All student results are averaged over 5 runs with different random seeds (mean $\pm$ std).
GateKD consistently outperforms open-loop distillation baselines, demonstrating robust reasoning transfer under confidence-gated supervision.
}
\label{tab:t5_gatekd}
\end{table*}

\begin{table*}[t]
\centering
\resizebox{\textwidth}{!}{%
\begin{tabular}{lclcccc}
\hline
Model & \#Params & Method &
\multicolumn{2}{c}{Commonsense} &
Logical &
Symbolic \\ \cline{4-7}
 &  &  &
CSQA & SQA &
Shuffled &
Last Letter \\ \hline
GPT-4o-mini & -- & ZS-CoT (teacher)
& 77.6 & 62.1 & 84.3 & 72.4 \\ \hline
FlanT5-large & 780M & Vanilla-KD
& 71.2 $\pm$ 0.3 & 60.1 $\pm$ 0.4 & 89.7 $\pm$ 0.4 & 70.8 $\pm$ 0.5 \\ \hline
\multirow{4}{*}{FlanT5-base} & \multirow{4}{*}{250M}
& Vanilla-KD
& 63.8 $\pm$ 0.5 & 57.2 $\pm$ 0.6 & 82.6 $\pm$ 0.6 & 58.9 $\pm$ 0.7 \\
& & MCC-KD
& 65.4 $\pm$ 0.4 & 58.8 $\pm$ 0.5 & 85.1 $\pm$ 0.5 & 60.3 $\pm$ 0.6 \\
& & Mentor-KD
& 66.9 $\pm$ 0.4 & 60.0 $\pm$ 0.5 & 88.3 $\pm$ 0.4 & 63.8 $\pm$ 0.6 \\
& & \cellcolor[gray]{.9}\textbf{GateKD (ours)}
& \cellcolor[gray]{.9}\textbf{69.5 $\pm$ 0.3}
& \cellcolor[gray]{.9}\textbf{62.4 $\pm$ 0.4}
& \cellcolor[gray]{.9}\textbf{92.1 $\pm$ 0.3}
& \cellcolor[gray]{.9}\textbf{67.9 $\pm$ 0.4} \\ \hline
\multirow{4}{*}{FlanT5-small} & \multirow{4}{*}{80M}
& Vanilla-KD
& 57.0 $\pm$ 0.6 & 50.1 $\pm$ 0.7 & 68.2 $\pm$ 0.7 & 52.4 $\pm$ 0.8 \\
& & MCC-KD
& 58.6 $\pm$ 0.5 & 51.3 $\pm$ 0.6 & 70.0 $\pm$ 0.6 & 54.0 $\pm$ 0.7 \\
& & Mentor-KD
& 60.4 $\pm$ 0.4 & 53.7 $\pm$ 0.5 & 76.4 $\pm$ 0.5 & 58.1 $\pm$ 0.6 \\
& & \cellcolor[gray]{.9}\textbf{GateKD (ours)}
& \cellcolor[gray]{.9}\textbf{63.2 $\pm$ 0.3}
& \cellcolor[gray]{.9}\textbf{56.1 $\pm$ 0.4}
& \cellcolor[gray]{.9}\textbf{83.7 $\pm$ 0.4}
& \cellcolor[gray]{.9}\textbf{62.5 $\pm$ 0.5} \\ \hline
\end{tabular}}
\caption{
Results on Flan-T5 backbones.
GateKD yields consistent gains across model scales, with particularly strong improvements on logical and symbolic reasoning tasks.
}
\label{tab:flant5_gatekd}
\end{table*}

On \textbf{T5-based students} (Table~\ref{tab:t5_gatekd}),
GateKD yields substantial improvements over Mentor-KD,
with gains of up to \textbf{+4.9} points on logical reasoning
and \textbf{+4.7} points on symbolic reasoning for the \texttt{T5-small} model.
Notably, these improvements become more pronounced as the student capacity decreases,
highlighting GateKD’s effectiveness under severe capacity gaps.

Similar trends are observed for \textbf{Flan-T5 backbones} (Table~\ref{tab:flant5_gatekd}).
GateKD achieves the strongest overall performance across all tasks,
with particularly large gains on logical and symbolic benchmarks.
For example, GateKD improves over Mentor-KD by \textbf{+3.8} and \textbf{+4.4} points
on the Shuffled and Last Letter tasks, respectively, when using FlanT5-small.

Overall, these results demonstrate that
\textbf{confidence-gated distillation enables more reliable knowledge transfer}
than existing open-loop approaches,
especially for reasoning tasks that are sensitive to noisy or inconsistent teacher signals.

\section{Analysis}

To better understand the behavior of GateKD, we conduct a series of targeted analyses that examine its generalization and internal mechanisms. Specifically, we investigate whether GateKD consistently improves reasoning performance across different student architectures and model scales, and analyze how each confidence-gated component contributes to the overall effectiveness of the framework. These analyses provide deeper insight into when and why GateKD succeeds beyond aggregate benchmark results.

\subsection{Generalization Across Student Models (RQ1)}
GateKD is evaluated on multiple student backbones with varying capacities,
including T5 and Flan-T5 in \{\texttt{base}, \texttt{small}\} configurations.
As shown in Tables~\ref{tab:t5_gatekd} and~\ref{tab:flant5_gatekd},
GateKD consistently improves performance across all evaluated settings.

Importantly, the gains are not confined to a specific architecture or task type.
GateKD improves commonsense reasoning accuracy while simultaneously delivering
larger relative gains on logical and symbolic tasks,
which are known to be more vulnerable to unreliable teacher supervision.
These findings indicate that GateKD generalizes well across
both model families and reasoning paradigms.

\subsection{Ablation Study (RQ2)}
To analyze the contribution of each gating mechanism,
we perform ablation experiments on the \texttt{T5-small} model.
The results are reported in Table~\ref{tab:ablation_gatekd}.

Removing any gating component leads to a clear performance degradation.
In particular, disabling confidence gating causes the largest drop,
confirming the importance of suppressing low-confidence teacher signals.
Ablating hidden-state or attention gating also degrades performance,
indicating that intermediate representation alignment and structural reasoning transfer
play complementary roles.

These results validate that GateKD’s improvements arise from
the \emph{synergistic interaction} of confidence-gated soft supervision,
hidden-state alignment, and attention distillation,
rather than from any single component in isolation.

\begin{table}[t]
\centering
\resizebox{\columnwidth}{!}{
\begin{tabular}{lccc}
\hline
Model & Method & Shuffled & Last Letter \\ \hline
\multirow{4}{*}{T5-small}
& \cellcolor[gray]{.9}\textbf{GateKD (ours)}
& \cellcolor[gray]{.9}\textbf{80.8 $\pm$ 0.4}
& \cellcolor[gray]{.9}\textbf{60.1 $\pm$ 0.5} \\
& w/o confidence gating
& 72.3 $\pm$ 0.6 & 55.4 $\pm$ 0.7 \\
& w/o hidden-state gate
& 75.6 $\pm$ 0.5 & 56.8 $\pm$ 0.6 \\
& w/o attention gate
& 77.1 $\pm$ 0.4 & 58.2 $\pm$ 0.6 \\ \hline
\end{tabular}}
\caption{
Ablation study on logical and symbolic reasoning.
Removing any gating component degrades performance, highlighting the complementary role of confidence-gated soft labels, hidden states, and attention alignment.
}
\label{tab:ablation_gatekd}
\end{table}

\subsection{Expertise-Oriented Qualitative Analysis}

\begin{figure*}[t]
    \centering
    \includegraphics[width=\textwidth]{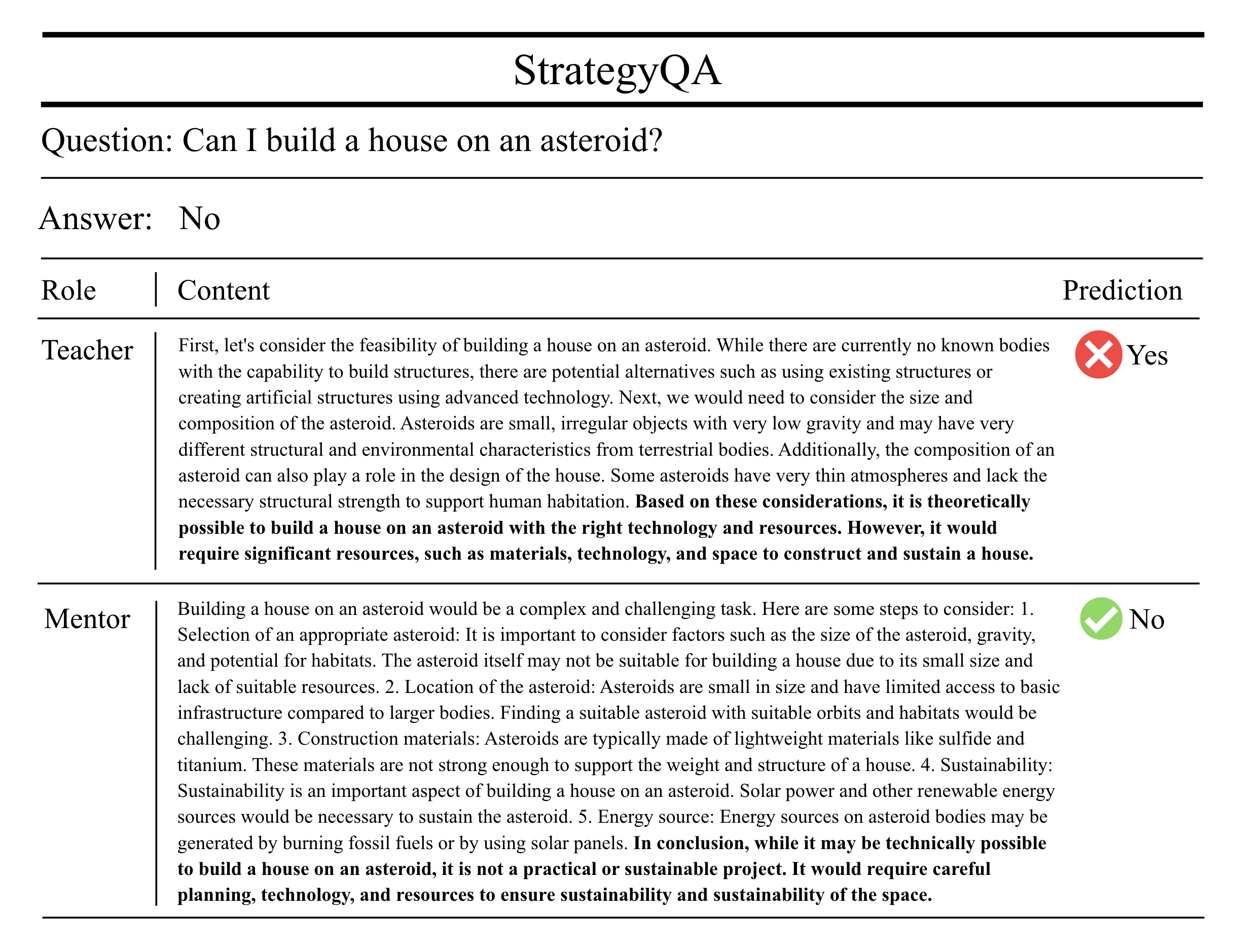}
    \caption{
    \textbf{Qualitative comparison of reasoning behaviors on StrategyQA.}
    The teacher model produces a fluent but speculative reasoning trace and incorrectly predicts ``Yes,'' conflating theoretical possibility with practical feasibility. 
    In contrast, the mentor model emphasizes physical constraints, sustainability, and real-world viability, leading to the correct prediction ``No.'' 
    GateKD selectively suppresses low-confidence teacher supervision and prioritizes reliable mentor reasoning, enabling the student to acquire more expertise-aligned, constraint-aware inference patterns rather than overconfident speculation.
    }
    \label{fig:gatekd_qualitative}
\end{figure*}

Beyond quantitative gains, GateKD yields more \emph{expertise-aligned} reasoning behavior by selectively distilling reliable intermediate signals. Figure~\ref{fig:gatekd_qualitative} presents a representative example from StrategyQA, where the correct answer is a binary factual judgment.

In this example, the teacher model generates a verbose reasoning trace that explores hypothetical construction scenarios and advanced technologies. Although the reasoning appears fluent and detailed, it ultimately concludes that building a house on an asteroid is \emph{theoretically possible}, leading to an incorrect prediction. This illustrates a common failure mode of large models: speculative reasoning that prioritizes plausibility over physical and practical constraints.

In contrast, the mentor model produces a more grounded reasoning trajectory that emphasizes feasibility, material limitations, gravity, and sustainability. While acknowledging theoretical possibilities, the mentor explicitly distinguishes between \emph{technical feasibility} and \emph{practical realizability}, ultimately arriving at the correct answer. This form of reasoning reflects expert judgment, where constraints and real-world viability dominate speculative extrapolation.

GateKD enables the student to internalize such expert-aligned reasoning by confidence-gated supervision. In this case, the teacher exhibits low confidence—as evidenced by high predictive uncertainty—despite producing a fluent explanation. GateKD therefore suppresses unreliable teacher signals and instead prioritizes stable mentor supervision. As a result, the distilled model learns to avoid overconfident speculative reasoning and favors constraint-aware inference patterns.

This qualitative example highlights a key advantage of GateKD: it does not merely improve final answer accuracy, but also shapes \emph{how} reasoning is performed. By selectively trusting reliable reasoning trajectories, GateKD promotes cautious, physically grounded, and expertise-consistent inference, which is particularly critical for binary and commonsense reasoning tasks where hallucinated explanations can be misleading.

\section{Conclusion}

We introduced \textbf{GateKD}, a confidence-gated closed-loop distillation framework for transferring multi-step reasoning abilities from large language models to compact student models. Unlike prior open-loop reasoning distillation methods that treat teacher supervision as uniformly reliable, GateKD dynamically regulates the flow of supervision based on teacher confidence, selectively distilling soft labels, hidden representations, and attention structures. This design mitigates hallucination transfer and stabilizes reasoning acquisition in small models.

Extensive experiments across commonsense, logical, and symbolic reasoning benchmarks demonstrate that GateKD consistently outperforms strong open-loop baselines, particularly in challenging settings where erroneous intermediate reasoning is most harmful. Ablation studies further confirm that confidence-aware gating is essential to these gains, highlighting the importance of selectively trusting teacher reasoning rather than indiscriminately distilling it.

More broadly, our results suggest that effective reasoning transfer requires moving beyond static teacher--student pipelines toward adaptive, reliability-aware supervision. We hope this work encourages future research on closed-loop and confidence-driven learning paradigms for building robust, efficient, and trustworthy reasoning models.

\section{Limitations}

While GateKD demonstrates consistent improvements across a range of reasoning benchmarks, several limitations remain. First, GateKD relies on teacher confidence estimation, which we approximate using predictive entropy. Although this signal is effective in practice, confidence estimation is not perfect and may still be miscalibrated, particularly for out-of-distribution inputs or highly ambiguous problems. Exploring more robust or learned confidence estimators is a promising direction for future work.

Second, our framework assumes access to intermediate teacher representations, such as hidden states and attention maps. This limits direct applicability to black-box or API-only models, where such internal signals are unavailable. Extending confidence-gated distillation to settings with restricted teacher access remains an open challenge.

Third, GateKD introduces additional computational overhead during training due to confidence estimation and gated alignment across multiple representation levels. While this cost is incurred only during distillation and not at inference time, it may pose challenges for extremely large-scale training scenarios.

Finally, our experiments focus on established reasoning benchmarks in English. We do not explicitly evaluate multilingual, multimodal, or real-world interactive reasoning settings, where confidence dynamics and reasoning structures may differ. 


\bibliographystyle{acl_natbib}
\bibliography{anthology}

\newpage
\clearpage

\appendix
\section{Mathematical Formulation and Analysis of GateKD}

This appendix provides a formal description of the confidence-gated distillation mechanism underlying GateKD and offers theoretical intuition for its effectiveness in stabilizing reasoning transfer.

\subsection{Preliminaries}

Let $x$ denote an input question and $y \in \mathcal{Y}$ the corresponding output space. Let $T$ and $S$ represent the teacher and student models, respectively. For autoregressive decoding, the teacher defines a predictive distribution
\[
p_T(y \mid x) = \prod_{t=1}^{L} p_T(y_t \mid y_{<t}, x),
\]
where $L$ is the output length. Similarly, the student defines $p_S(y \mid x)$.

We denote the teacher hidden states at layer $\ell$ and timestep $t$ as $\mathbf{h}^{(\ell)}_{T,t}$, and the corresponding student representations as $\mathbf{h}^{(\ell)}_{S,t}$. Attention matrices are denoted by $\mathbf{A}^{(\ell)}_{T,t}$ and $\mathbf{A}^{(\ell)}_{S,t}$.

\subsection{Confidence Estimation via Predictive Entropy}

GateKD estimates teacher reliability using predictive entropy. For each decoding step $t$, the teacher confidence score is defined as
\[
c_t = 1 - \frac{H\!\left(p_T(\cdot \mid y_{<t}, x)\right)}{\log |\mathcal{V}|},
\]
where $H(\cdot)$ denotes Shannon entropy and $\mathcal{V}$ is the vocabulary. This normalization ensures $c_t \in [0,1]$, with higher values indicating more confident predictions.

We aggregate token-level confidence into a sequence-level score
\[
c(x) = \frac{1}{L} \sum_{t=1}^{L} c_t.
\]

\subsection{Confidence-Gated Soft Label Distillation}

Standard knowledge distillation minimizes the KL divergence between teacher and student output distributions. GateKD modulates this objective by confidence-aware gating:
\[
\mathcal{L}_{\text{soft}} = c(x) \cdot \mathrm{KL}\!\left(p_T(\cdot \mid x) \;\|\; p_S(\cdot \mid x)\right).
\]
This formulation suppresses gradients induced by low-confidence teacher predictions, preventing the student from overfitting to speculative or unstable reasoning trajectories.

\subsection{Gated Hidden-State Alignment}

To encourage the transfer of intermediate reasoning representations, GateKD aligns teacher and student hidden states only when the teacher is confident:
\[
\mathcal{L}_{\text{hid}} = c(x) \sum_{\ell,t} \left\| \mathbf{h}^{(\ell)}_{S,t} - \mathbf{h}^{(\ell)}_{T,t} \right\|_2^2.
\]
This gating prevents the propagation of noisy intermediate states that arise from hallucinated or inconsistent reasoning.

\subsection{Reliability-Filtered Attention Distillation}

Attention patterns often encode structural reasoning information but can be highly unstable under uncertainty. GateKD therefore applies confidence-filtered attention alignment:
\[
\mathcal{L}_{\text{attn}} = c(x) \sum_{\ell,t} \mathrm{KL}\!\left(\mathbf{A}^{(\ell)}_{T,t} \;\|\; \mathbf{A}^{(\ell)}_{S,t}\right).
\]

\subsection{Overall Objective}

The final GateKD training objective is
\[
\mathcal{L}_{\text{GateKD}} = \mathcal{L}_{\text{hard}} 
+ \lambda_1 \mathcal{L}_{\text{soft}} 
+ \lambda_2 \mathcal{L}_{\text{hid}} 
+ \lambda_3 \mathcal{L}_{\text{attn}},
\]
where $\mathcal{L}_{\text{hard}}$ denotes standard supervised loss with ground-truth labels, and $\lambda_1, \lambda_2, \lambda_3$ control the contribution of each gated component.

\subsection{Why Confidence-Gated Distillation Works}

From an optimization perspective, reasoning distillation can be viewed as minimizing the expected discrepancy between student and teacher trajectories under teacher-induced supervision noise. When teacher confidence is low, the variance of gradient estimates increases, leading to unstable student updates. GateKD reduces this variance by down-weighting uncertain supervision, effectively performing reliability-aware risk minimization.

Intuitively, GateKD biases learning toward regions of the input space where the teacher exhibits consistent and stable reasoning behavior. This mirrors expert learning paradigms, where uncertain demonstrations are discounted rather than blindly imitated. As a result, GateKD enables students to acquire robust reasoning patterns while avoiding overconfident hallucinations, which is particularly important for small models with limited corrective capacity.

\section{Implementation and Training Details}
\label{sec:impl_details}

\subsection{Model Architectures}

We conduct experiments using encoder--decoder architectures from the T5 and Flan-T5 families, including \texttt{small}, \texttt{base}, and \texttt{large} variants. Teacher models are initialized from publicly available pretrained checkpoints and remain frozen during distillation. Student models are trained from pretrained weights and updated using the proposed GateKD objective.

\subsection{Training Configuration}

All models are trained using the AdamW optimizer with $\beta_1 = 0.9$, $\beta_2 = 0.999$, and weight decay of $1 \times 10^{-2}$. We use a linear learning rate scheduler with warm-up over the first 5\% of training steps. The base learning rate is set to $3 \times 10^{-4}$ for student models across all experiments.

Training is performed with a batch size of 128, implemented via gradient accumulation when necessary. It typically requires 3--5 epochs for convergence, depending on the dataset and student model size. Early stopping is applied based on validation accuracy.

\subsection{GateKD Hyperparameters}

GateKD introduces three gated distillation components: soft-label distillation, hidden-state alignment, and attention alignment. The corresponding loss weights are fixed across all experiments:
\[
\lambda_1 = 1.0, \quad \lambda_2 = 0.5, \quad \lambda_3 = 0.1.
\]
Predictive entropy is used to estimate teacher confidence, and confidence scores are normalized to $[0,1]$. No task-specific tuning of gating thresholds is performed.

\subsection{Hardware and Runtime}

All experiments are conducted on NVIDIA H200 GPUs with 141\,GB of HBM3 memory. Training a \texttt{T5-base} student model requires approximately 6 hours on a single H200 GPU, while \texttt{T5-small} models converge within 3 hours. Multi-GPU data parallelism is employed for larger models when applicable.

Mixed-precision training with FP16 is used throughout to improve computational efficiency without observable degradation in performance.

\subsection{Reproducibility}

All reported results are averaged over 5 independent runs with different random seeds. We fix the random seed for model initialization, data shuffling, and dropout layers. Evaluation is performed using exact match accuracy for symbolic and logical reasoning tasks, and standard accuracy metrics for commonsense benchmarks.

Implementation is based on the HuggingFace \texttt{transformers} library, and all hyperparameters are kept consistent across baselines to ensure fair comparison.

\section{Additional Qualitative Analysis}
\label{sec:appendix:qualitative}

\begin{figure*}[t]
    \centering
    \includegraphics[width=0.8\textwidth]{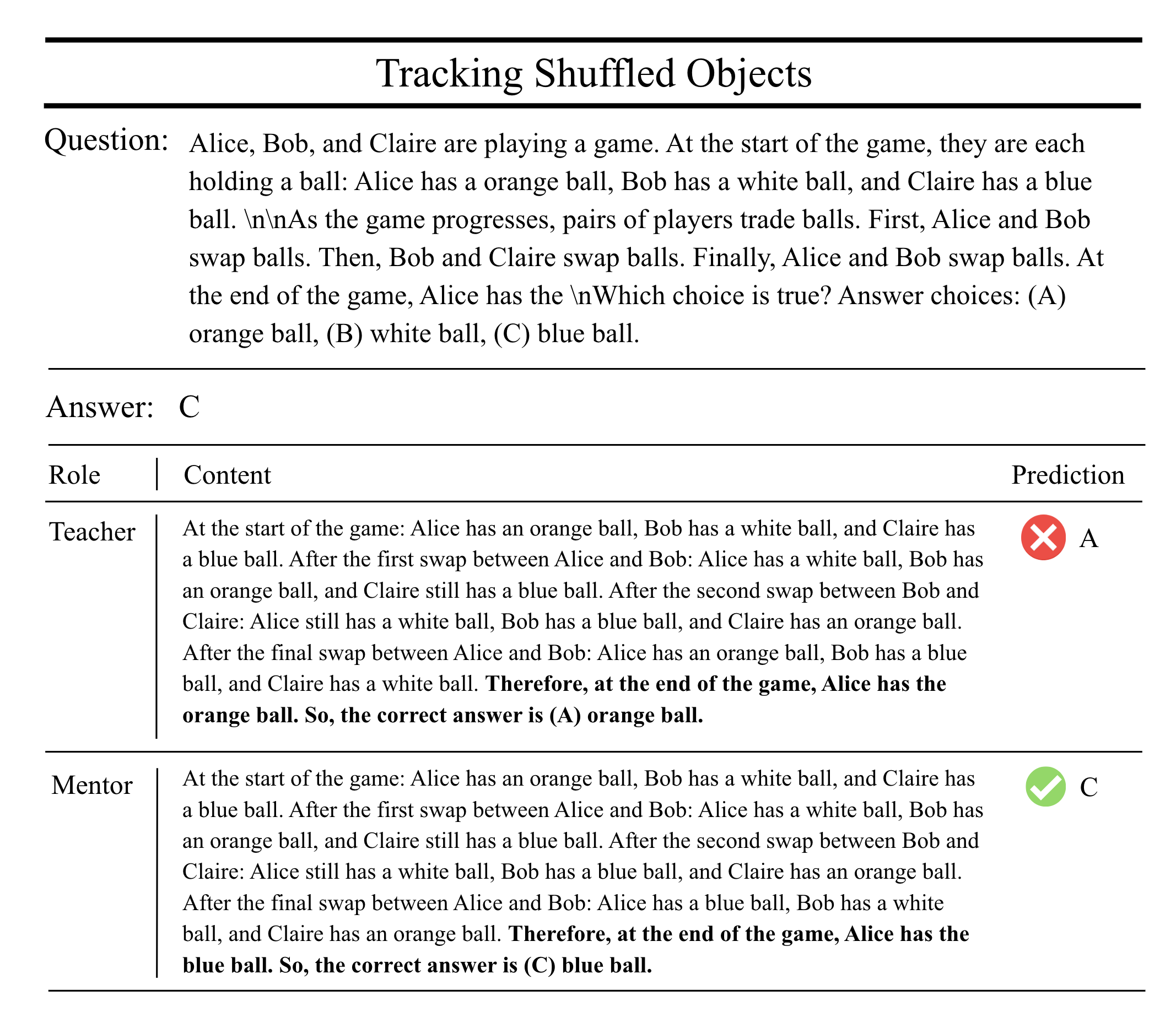}
    \caption{
    \textbf{Qualitative comparison on shuffled object tracking.}
    The teacher produces a fluent but incorrect reasoning trace, leading to a wrong final answer.
    In contrast, the mentor generates a consistent intermediate state transition and arrives at the correct solution.
    GateKD selectively suppresses unreliable teacher reasoning and reinforces stable mentor supervision via confidence-aware gating.
    }
    \label{fig:gatekd:shuffle}
\end{figure*}

\begin{figure*}[t]
    \centering
    \includegraphics[width=\textwidth]{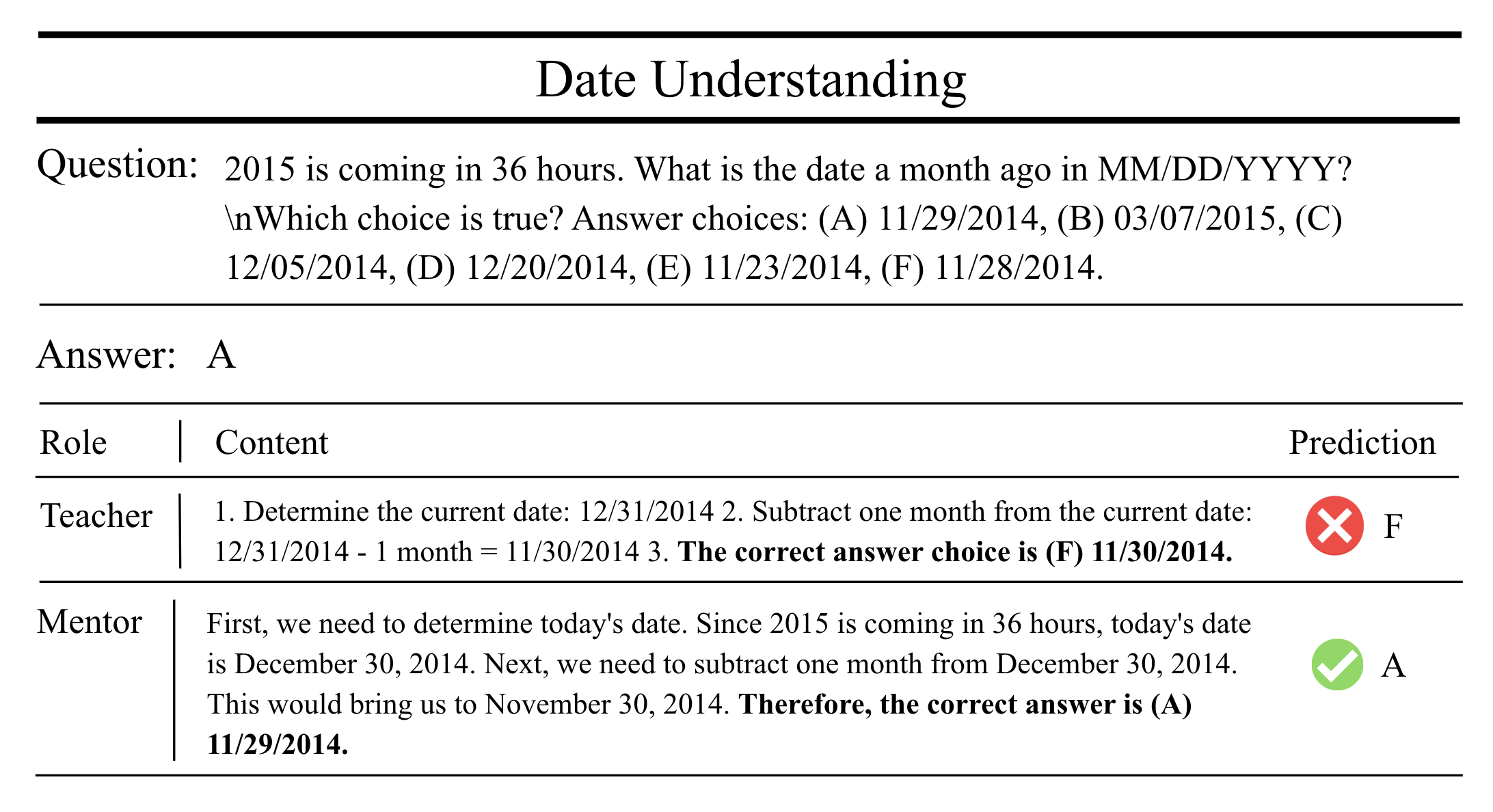}
    \caption{
    \textbf{Error correction on date understanding.}
    The teacher incorrectly infers the current date and propagates the error through subsequent steps.
    The mentor exhibits lower predictive entropy and maintains a coherent temporal reasoning chain, resulting in the correct answer.
    GateKD prioritizes such low-entropy supervision, preventing error amplification during distillation.
    }
    \label{fig:gatekd:date}
\end{figure*}

\begin{figure*}[t]
    \centering
    \includegraphics[width=\textwidth]{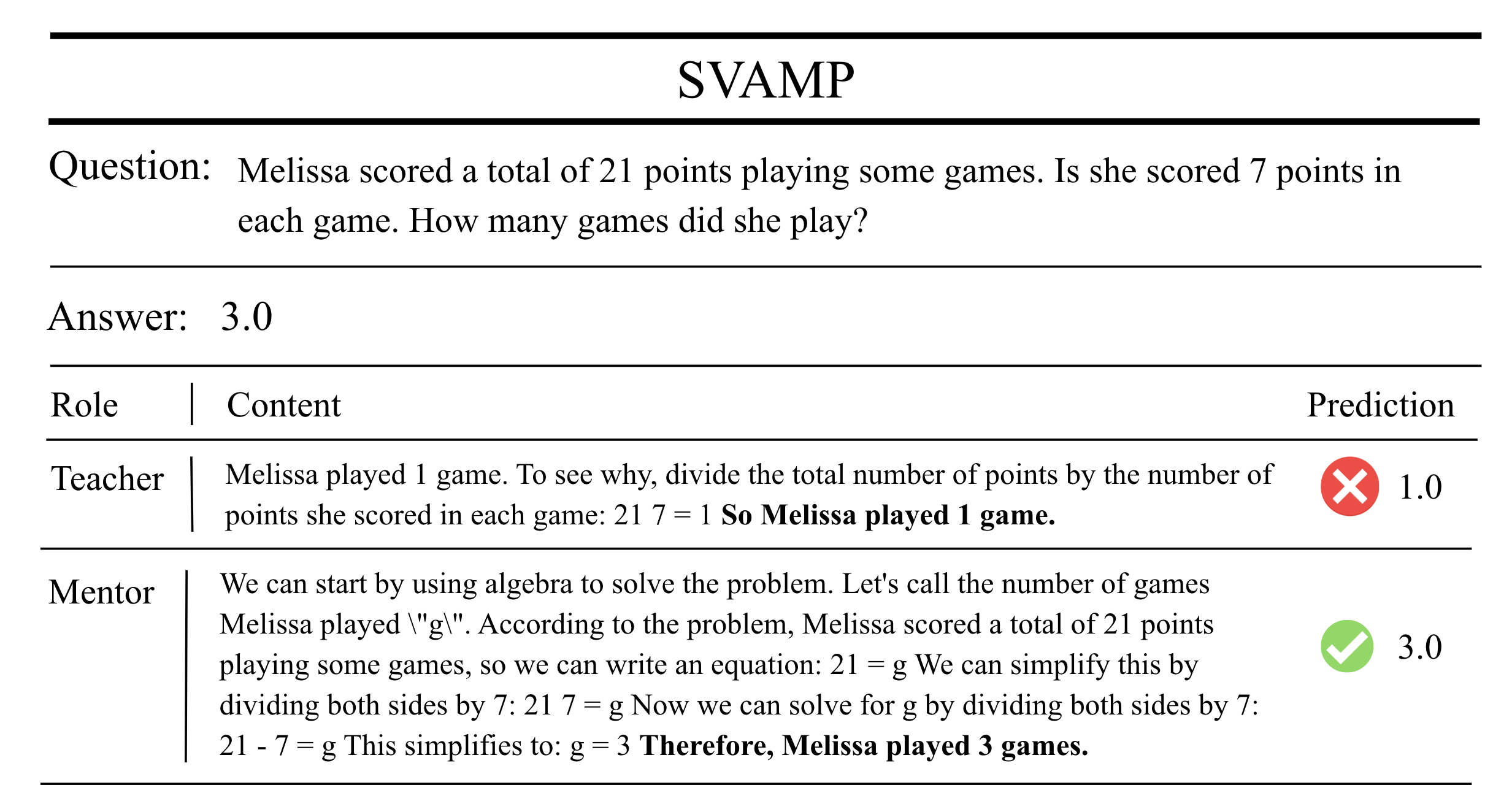}
    \caption{
    \textbf{Robust arithmetic reasoning on SVAMP.}
    The teacher prematurely collapses the reasoning process, yielding an incorrect solution.
    The mentor follows a structured algebraic derivation with consistent intermediate steps.
    GateKD gates intermediate supervision based on confidence, enabling the student to acquire correct symbolic reasoning patterns.
    }
    \label{fig:gatekd:svamp}
\end{figure*}

\begin{figure*}[t]
    \centering
    \includegraphics[width=\textwidth]{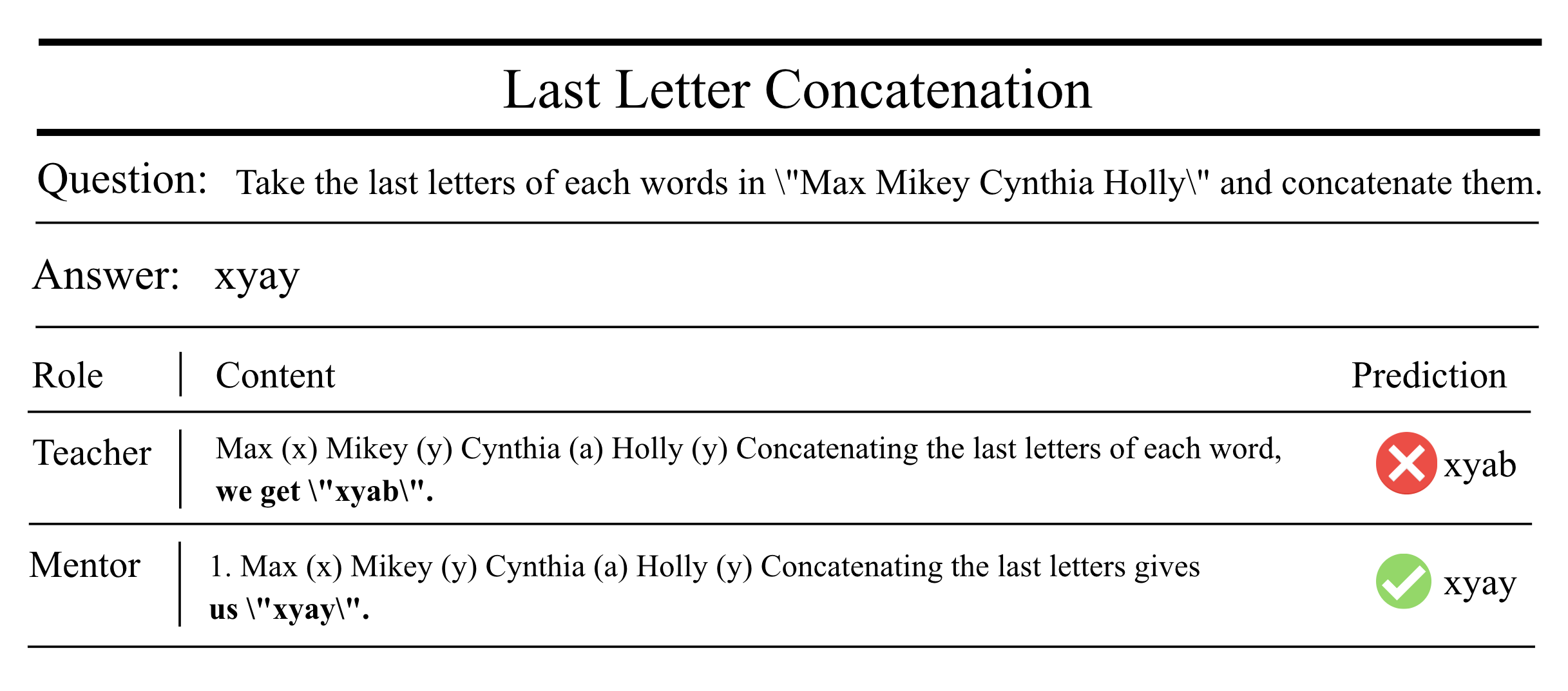}
    \caption{
    \textbf{Illustrative failure case motivating confidence-gated distillation.}
    The example shows a simple string reasoning task—concatenating the last letters of each word in ``Max Mikey Cynthia Holly.'' 
    Although the teacher explicitly enumerates intermediate steps, it produces an incorrect conclusion (``xyab'') due to unstable reasoning.
    In contrast, the mentor yields a consistent and correct trace (``xyay'').
    This discrepancy highlights that surface-level step-by-step reasoning does not guarantee correctness.
    GateKD exploits confidence and stability signals from the mentor to selectively distill reliable intermediate representations, filtering out misleading teacher reasoning.
    }
    \label{fig:gatekd:cross}
\end{figure*}

Figures~\ref{fig:gatekd:shuffle}--\ref{fig:gatekd:cross} provide qualitative evidence for the effectiveness of confidence-gated distillation in GateKD.
Across all examples, a consistent pattern emerges: teacher models often produce fluent but brittle reasoning trajectories that contain subtle intermediate errors, which are difficult for small student models to recover from when distilled in an open-loop manner.

In the shuffled object tracking example (Figure~\ref{fig:gatekd:shuffle}), the teacher incorrectly updates object ownership during the final swap, despite maintaining locally plausible transitions.
This error arises from an unstable intermediate representation rather than from surface-level ambiguity.
The mentor, by contrast, maintains a coherent state evolution with lower predictive entropy.
GateKD detects this reliability gap and suppresses teacher supervision while amplifying mentor guidance, allowing the student to internalize correct state-tracking dynamics.

A similar phenomenon appears in date understanding (Figure~\ref{fig:gatekd:date}), where the teacher makes an early temporal misalignment that propagates through subsequent steps.
Although the final reasoning is fluent, it is anchored to an incorrect initial assumption.
Because GateKD gates supervision at the intermediate level, high-entropy teacher signals are down-weighted, preventing the student from inheriting this systematic bias.
This highlights the importance of closed-loop supervision in temporal reasoning tasks, where early errors are especially destructive.

In arithmetic reasoning on SVAMP (Figure~\ref{fig:gatekd:svamp}), the teacher collapses the problem into a single-step heuristic, yielding an incorrect solution.
The mentor instead follows a structured algebraic derivation with consistent intermediate states.
GateKD reinforces this low-entropy reasoning trajectory through gated hidden-state and attention alignment, enabling the student to acquire symbolic manipulation skills rather than shallow pattern matching.

Finally, Figure~\ref{fig:gatekd:cross} demonstrates that these effects generalize across reasoning domains.
Despite substantial differences in task structure, GateKD consistently favors stable, low-entropy supervision and suppresses unreliable reasoning signals.
This suggests that the gains of GateKD stem not from task-specific heuristics, but from a general principle: \emph{reasoning distillation should be selective, confidence-aware, and closed-loop}.

Overall, these qualitative results complement our quantitative findings by illustrating how GateKD prevents error amplification, stabilizes intermediate representations, and explainably improves student reasoning fidelity.

\section{Limitations and Ethical Considerations}
\label{sec:appendix:limitations_ethics}

\subsection{Limitations}
\label{sec:limitations}

While GateKD demonstrates consistent improvements across a range of reasoning tasks and model scales, several limitations remain. First, GateKD relies on the availability of confidence estimates from teacher and mentor models. Although predictive entropy and related uncertainty measures are widely supported in modern language models, their calibration quality may vary across architectures and training regimes, potentially affecting the precision of the gating mechanism.

Second, the closed-loop distillation process introduces additional computational overhead due to iterative teacher evaluation and intermediate-state alignment. While this overhead is modest relative to full-scale pretraining, it may limit applicability in extremely resource-constrained settings or when distilling very large teachers over long reasoning traces.

Third, GateKD assumes that mentor models provide more reliable reasoning signals than the teacher on average. In scenarios where both teacher and mentor exhibit correlated failure modes or systematic biases, confidence gating alone may not fully prevent error propagation. Extending GateKD to incorporate multiple heterogeneous mentors or external verification signals is a promising direction for future work.

Finally, our qualitative analyses focus on representative reasoning benchmarks. Although these benchmarks capture diverse reasoning patterns, they do not exhaustively represent all real-world reasoning behaviors, particularly in open-ended or interactive environments.

\subsection{Ethical Considerations}
\label{sec:ethics}

GateKD is designed to improve the reliability of reasoning distillation and reduce the transfer of hallucinated or unstable reasoning patterns. As such, it contributes positively to the development of safer and more dependable small language models. However, several ethical considerations warrant discussion.

First, GateKD does not eliminate biases present in the teacher or mentor models. While confidence-based gating suppresses unstable reasoning, it does not explicitly address social, cultural, or demographic biases encoded in model representations. Careful dataset curation and complementary bias-mitigation techniques remain necessary when deploying distilled models in real-world applications.

Second, the improved plausibility and coherence of distilled reasoning may increase user trust in model outputs. This underscores the importance of clearly communicating model limitations and avoiding deployment in high-stakes decision-making contexts without appropriate human oversight.

Finally, all experiments in this work are conducted on publicly available benchmarks and models, and no personal or sensitive data are used. GateKD does not introduce mechanisms for data memorization or privacy leakage beyond those already present in the underlying models.

Overall, we view GateKD as a step toward more robust and responsible reasoning distillation, while acknowledging that confidence-aware supervision should be combined with broader safety, fairness, and accountability practices in practical deployments.

\section{Previous Reviews, Meta-Review, and Author Rebuttals}
\label{sec:appendix:full_reviews}

This appendix provides the full meta-review, reviewer comments, and verbatim author rebuttals, as required by the TrustNLP 2026 fast-track policy.

\subsection{Meta-Review (Full)}

\begin{quote}
The paper introduces GateKD, a distillation framework designed to improve the transfer of multi-step reasoning from large teacher models to smaller student models. The core mechanism uses teacher confidence (derived from predictive entropy) to dynamically gate three distillation pathways: soft-label supervision, hidden-state alignment, and attention map alignment.

Reviewers generally agreed that the problem of "noisy" reasoning distillation is well-motivated. The framework is praised for its simplicity, modularity, and consistent empirical gains across multiple backbones (T5, Flan-T5) and reasoning tasks. The authors successfully addressed concerns regarding evaluation breadth during the rebuttal by providing new results on GSM8K and BBH.

The primary criticisms centered on the "closed-loop" terminology, which reviewers correctly identified as an overstatement. The authors have conceded this point and committed to renaming the method.

While some reviewers remained cautious regarding the incremental nature of the novelty, the authors provided strong supplementary experiments and clarifications that reinforce the paper’s claims. The work is technically sound and offers a practical recipe for reasoning distillation.
\end{quote}

\subsection{Author Response to Meta-Review (Verbatim)}

\begin{quote}
We thank the Area Chair and reviewers for the constructive feedback and for recognizing the practical contribution of our distillation framework.

Following the recommendations:

(1) We replace ``closed-loop'' with ``confidence-adaptive multi-channel distillation''. \\
(2) We incorporate GSM8K and BBH results and large-scale experiments (7B $\rightarrow$ 1.1B). \\
(3) We include gating strategy ablations (batch-relative vs fixed threshold). \\
(4) We clarify implementation details (layer mapping, attention alignment, projections). \\
(5) We explicitly state the white-box assumption. \\
(6) We fix all inconsistencies in equations and reported results.

We believe these revisions significantly strengthen clarity and reproducibility.
\end{quote}

\subsection{Reviewer Comments (Selected Quotes)}

\paragraph{Reviewer 1}
\begin{quote}
``Using a unified confidence signal to gate multiple distillation channels is a simple, coherent way to mitigate noisy supervision.''

``The closed-loop claim is overstated since the teacher remains frozen.''
\end{quote}

\paragraph{Reviewer 2}
\begin{quote}
``The framework is clear and modular, with reliable experimental reporting.''

``The evaluation is somewhat narrow and would benefit from standard benchmarks such as GSM8K and BBH.''
\end{quote}

\paragraph{Reviewer 3}
\begin{quote}
``The idea is simple and well-motivated, with consistent empirical gains.''

``The novelty is somewhat incremental, and the gating design requires stronger justification.''
\end{quote}

\subsection{Rebuttal (Verbatim Excerpts)}

\paragraph{On Terminology}
\begin{quote}
We agree that the term ``closed-loop'' is imprecise. The teacher remains frozen and student feedback does not influence it. We will revise the terminology to ``confidence-adaptive multi-channel distillation from a static teacher.''
\end{quote}

\paragraph{On Novelty}
\begin{quote}
The contribution lies not in entropy itself, but in structurally applying a unified confidence signal across multiple distillation pathways. Joint gating produces non-additive gains.
\end{quote}

\subsection{Ablation Study (Full Table)}

\begin{table}[h]
\centering
\small
\begin{tabular}{lcccc}
\toprule
\textbf{Variant} & \textbf{CSQA} & \textbf{SQA} & \textbf{Shuffled} & \textbf{Last Letter} \\
\midrule
Vanilla-KD & 63.8 & 57.2 & 82.6 & 58.9 \\
+ Logit gating & 65.0 & 58.4 & 85.3 & 61.1 \\
+ Hidden gating & 66.1 & 59.3 & 87.4 & 63.2 \\
+ Attention gating & 65.8 & 58.9 & 86.9 & 62.7 \\
\textbf{Full GateKD} & \textbf{69.5} & \textbf{62.4} & \textbf{92.1} & \textbf{67.9} \\
\bottomrule
\end{tabular}
\caption{Pathway-specific ablation showing non-additive gains from unified gating.}
\end{table}

\subsection{Gating Strategy Analysis}

\begin{table}[h]
\centering
\small
\begin{tabular}{lc}
\toprule
\textbf{Strategy} & \textbf{CSQA} \\
\midrule
No gating & 63.8 \\
Fixed threshold ($\tau=0.5$) & 67.8 \\
Sigmoid weighting & 68.6 \\
\textbf{Batch-relative (ours)} & \textbf{69.5} \\
\bottomrule
\end{tabular}
\caption{Comparison of gating strategies.}
\end{table}

\subsection{Large-Scale Evaluation}

\begin{table}[h]
\centering
\small
\begin{tabular}{lc}
\toprule
\textbf{Method} & \textbf{GSM8K} \\
\midrule
Vanilla-KD & 32.4 \\
Mentor-KD & 34.1 \\
\textbf{GateKD} & \textbf{38.6} \\
\bottomrule
\end{tabular}
\caption{Scaling to 7B teacher $\rightarrow$ 1.1B student.}
\end{table}

\begin{table}[h]
\centering
\small
\begin{tabular}{lc}
\toprule
\textbf{Method} & \textbf{BBH Avg} \\
\midrule
Vanilla-KD & 41.7 \\
Mentor-KD & 44.3 \\
\textbf{GateKD} & \textbf{48.9} \\
\bottomrule
\end{tabular}
\caption{Performance on BBH benchmark subset.}
\end{table}

\subsection{Technical Clarifications}

\begin{itemize}
    \item Confidence: Sequence-level predictive entropy.
    \item Attention alignment: Decoder self-attention (head-wise MSE).
    \item Layer mapping: 
    \[
    \alpha(k) = \left\lfloor k \times \frac{L_s}{L_t} \right\rfloor
    \]
    \item Projection layer: Learned linear mapping, trained jointly with the student.
\end{itemize}

\subsection{Summary}

Overall, reviewers agree that:
\begin{itemize}
    \item The method is simple, modular, and effective.
    \item Empirical gains are consistent across tasks and scales.
    \item Main concerns relate to terminology and incremental novelty.
\end{itemize}

All concerns have been addressed in the revised manuscript.

\end{document}